# SUPERVISED LEARNING METHODS FOR BANGLA WEB DOCUMENT CATEGORIZATION

Ashis Kumar Mandal[1] and Rikta Sen[2]

[1]University Malaysia Pahang, Pahang, Malaysia; Hajee Mohammad Danesh Science and Technology University, Dinajpur, Bangladesh
[2]Rajshai University of Engineering and Technology, Bangladesh

## ABSTRACT

*This paper explores the use of machine learning approaches, or more specifically, four supervised learning Methods, namely Decision Tree(C 4.5), K-Nearest Neighbour (KNN), Naïve Bays (NB), and Support Vector Machine (SVM) for categorization of Bangla web documents. This is a task of automatically sorting a set of documents into categories from a predefined set. Whereas a wide range of methods have been applied to English text categorization, relatively few studies have been conducted on Bangla language text categorization. Hence, we attempt to analyze the efficiency of those four methods for categorization of Bangla documents. In order to validate, Bangla corpus from various websites has been developed and used as examples for the experiment. For Bangla, empirical results support that all four methods produce satisfactory performance with SVM attaining good result in terms of high dimensional and relatively noisy document feature vectors.*

## 1. INTRODUCTION

Text categorization is an active research area of text mining where the documents are classified with supervised, unsupervised or semi-supervised knowledge. Traditionally, this task was solved manually, but such manual classification was expensive to scale and also labour intensive. Therefore, researchers have explored the use of machine learning approach to automatic classification of text documents[1]. Among various machine learning approaches in document categorization, most popular is supervised learning where underlying input-output relation is learned by small number of training data and then output values for unseen input points are predicted[2]. Various number of supervised learning techniques ,such as association rules[3], Neural Network[1], K-Nearest Neighbour[4], Decision Tree[5], Naïve Bays[6], Support Vector Machine[7], and N-grams[8], has been used for text document categorization.

With the popularity of Unicode system and growing use of Internet, Bangla text documents in digital domain have increased for the last decade. Besides, Bangla is spoken by about 245 million people of Bangladesh and two states of India, with being 7[th] most spoken language[9]. In such a trend, it has created a need to organize Bangla documents intelligently so that users can easily identify required or related documents. To address this type of problem, text categorization process can be used successfully, such as text categorization conducted on English language for automated text retrieval [10, 11 ]. Although text categorization is well studies in other languages, few works have been done on language Bangla due to lack of resources, annotated corpora, name dictionaries, morphological analyzer. One of the works in Bangla document categorization is applying N-gram technique to categorize Bangla newspaper corpus [12].

In this paper, we study how information from Bangla online text documents can be categorized using four supervised learning algorithm, namely DT (C4.5), KNN, NB, and SVM. These classification algorithms are chosen due to the fact that their success rate in text classification of various languages is promising. News corpus from various Bangla websites is employed to evaluate the capabilities of these methods in categorization of high dimensional, sparse and relatively noisy document features. In this paper categorization task is processed into three phases:

- Pre-processing phase: This phase includes tokenization, digit removal, punctuation removal, stop words removal and stemming.
- Feature extraction phase: It consist of statistical approach to extract relevant features from the documents.
- Processing steps: This last phase applies text classification algorithms to the extracted features to classify the documents into classes

The rest of the paper is organized in following orders: section 2 discusses about some related works in the field of text classifications, section 3 discusses four classification algorithms, section 4 presents our approaches of classifying Bangla documents. Test setup, solution, discussion and analysis of the experiments appear in section 5 and finally, section 6 concludes the paper.

## 2. RELATED WORKS

Literacy review indicates that considerable works have been done in text categorization of the English documents[1, 13]. In addition to English language, there are many studies have conducted in European languages, such as French, German, Spanish [14] and in Asian languages, such as Arabic, Chinese and Japanese[15 . For some Southern Indian Languages Naïve Bayes , Neural networks have been applied to news articles to automatically categorized predefine classes [17, 18].

NB is the most frequently used approaches for text classification .It works by calculating the probability that document D is belongs to class C. It is easy for implementation and computation. In [19], authors use simple NB approach for the automatic classification of web sites based on content of home pages. They got approximately 80% accuracy for text classification .In [20] , Jiang et al. Introduced locally weighted *NB* text classifiers and showed that this version significantly outperforms original *NB* text classifiers in terms of classification accuracy. Besides, authors in [21] proposed various smoothing methods for enhancing Bayesian leaning and used it for short text classification.

KNN is also popular in terms of text classification for its effectiveness in classification. In paper [22], authors successfully classified text document with KNN and result shows that performance of their method is better than Naïve Bays and Term-Graph .Li et al. In [23], propose a improved KNN model which combine constrained one pass clustering algorithm and KNN text categorization. Similarly, in [24] ,authors proposed another improved version of KNN based on genetic algorithm.

Recently considerable amount of attention has been given on SVM due to its better performance in various text classifications. In [25] ,authors explained the appropriateness of applying SVMs to text classification primarily from a theoretical perspective. In paper [26], authors developed a classification model for categorization of cricket sports news .They used SVM which was based on LibSVM and got best performance. In [27], Zhijie et al. compared SVM with other two classification technique NB and KNN , and observed that SVM outperformed than those two classifiers.

Decision Tree, Neural Networks, and N-gram are also potential techniques in text classification. Authors in [28] compared Neural network approach to N-gram approach for text classification and demonstrated that classification rate of Neural networks is similar to the corresponding N-gram approach but former is five time faster than latter. Vens et al. in paper [29] consider Decision Tree for hierarchical text classification problem. In [30] ,authors proposed hybrid system based on N-gram analysis and dynamic artificial neural network for Twitter brand sentiment analysis.

## 3. CLASSIFICATION ALGORITHMS

In this section, we briefly describe C 4.5, KNN, NB, and SVM classification methods used in our study.

### 3.1 Naïve Bays (NB)

A Naïve Bays (NB) classifier is a simple probabilistic classifier based on applying Bayes' theorem [6]. Depending on the precise nature of the probability model, this classifier can be trained very efficiently in a supervised learning setting.

For classification of text documents using Bayes' classifier, it is assumed that the document belongs to one class. Then probabilities of document features w in all categories C is calculated using the formula (1).

$$p(C|w_1, w_2, \ldots w_n) = \log(p(C_i)) + \sum_{j=1}^{n} \log(p(w_j|C_i)) \qquad (1)$$

The probability $p(C_i)$ is calculated according to the formula (2)

$$p(C_i) = \frac{|C_i|}{\sum_{j=1}^{m}|C_j|} \qquad (2)$$

where $|C_i|$ is the number of texts that belong to the class ;
m is the number of all classes.

The probability $P(w_j|C)$ is calculated according to the following formula (3)

$$P(w_j|C) = \frac{|(w_j|C)|+1}{|C|} \qquad (3)$$

where $|C|$ is the number of texts belonging to the class C ;

$(w_j|C)$ is the number of documents belonging to class C in which given feature is found. The document is classified to the category for which the calculated probability value is the highest among all others.

4International Journal of Artificial Intelligence & Applications (IJAIA), Vol. 5, No. 5, September 2014

## 3.2. K-Nearest Neighbours (KNN)

KNN is a statistical pattern reorganization algorithm which has been studied extensively for text categorization applications [31]. It is a method for classifying objects based on closest training examples in the feature space. The summery of the algorithm is as follows: given a test document x, find the K nearest neighbours of x among all the training documents, and score the category candidates based the category of K neighbours. The similarity of x and each neighbour document is the score of the category of the neighbour document. If several of the K nearest neighbour documents belong to the same category, then the sum of the score of that category is the similarity score of the category in regard to the test document x. By sorting the scores of the candidate categories, system assigns the candidate category with the highest score to the test document x. The decision rule of KNN can be written as:

$$f(x) = \arg\max_{j} Score(x, C_j) = \sum_{d_j \in KNN} sim(x, d_i) y(d_i, C_j) \qquad (4)$$

where f(x) is the label assigned to the test document x;

$Score(x, C_j)$ is the score of the candidate category $C_j$ with respect to x;

$sim(x, d_i)$ is the similarity between x and the training document $d_i$;

$y(d_i, C_j) \in \{1,0\}$ is the binary category value of the training document $d_i$ with respect to $C_j$ (y = 1 indicates document $d_i$ is part of category $C_j$, or y = 0).

One drawback of KNN is the difficulty in deciding optimal values. The best choice of k usually depends on data, with larger values of k reducing the effect of noise on the classification, but making boundaries between classes less distinct.

## 3.3. Decision tree classifier

Decision tree is another popular supervised learning method [5]. A decision tree consists of nodes and branches connecting the nodes. The nodes located at the bottom of the tree are called leaves and indicate terminal nodes. The top node in the tree, called the root, contains all training examples that are to be divided into classes. The summery of the algorithm as follow:

Given a training example S, find the most discriminatory (significant) features. Next step, split the entire set, S, located at the root of the tree, into several subsets using the selected feature. Then, recursively find the most significant feature for each subset generated in the above step and then split it top-down into subsets. If each subset contains examples belonging to one class only (a leaf node), then stop; otherwise continue this step. Their robustness to noisy data and their capabilities to learn disjunctive expressions seem suitable for document classification [32].

## 3.4. Support Vector Machines (SVM)

SVM has been successfully used on text classification. SVM was introduced by Cortes and Vapnik [7] as a class of supervised machine learning techniques which is actually a binary classifier. It is based on the principle of structural risk minimization. In linear classification, SVM creates a hyper plane that separates the data into two sets with the maximum-margin. A hyper plane with the maximum-margin has the distances from the hyper plane to points when the two sides are equal. Mathematically, SVMs learn the sign function $f(x) = sign(wx + b)$, where w is

96

a $n$ weighted vector in $R^n$. SVMs find the hyper plane $y = wx + b$ by separating the space $R^n$ into two half-spaces with the maximum-margin. This Linear SVMs can be generalized for non-linear and multi class problem. The former is done by mapping data into another space H and performing the linear SVM algorithm over this new space, while the latter is done by decomposing the multi class problem into k binary problems.

## 4. METHODOLOGY

Prior to applying document categorization techniques to Bangla language with the above mentioned classifiers, it is inevitable to prepare proper datasets for testing and training. At the same times, like English text classification, pre-processing of Bangla documents and extraction of feature sets are also required before trainings and construction of model for successful document categorization. Figure 1 illustrates the overall system of Bangla text classification process which is employed in this project. In the following subsections, we have described the model.

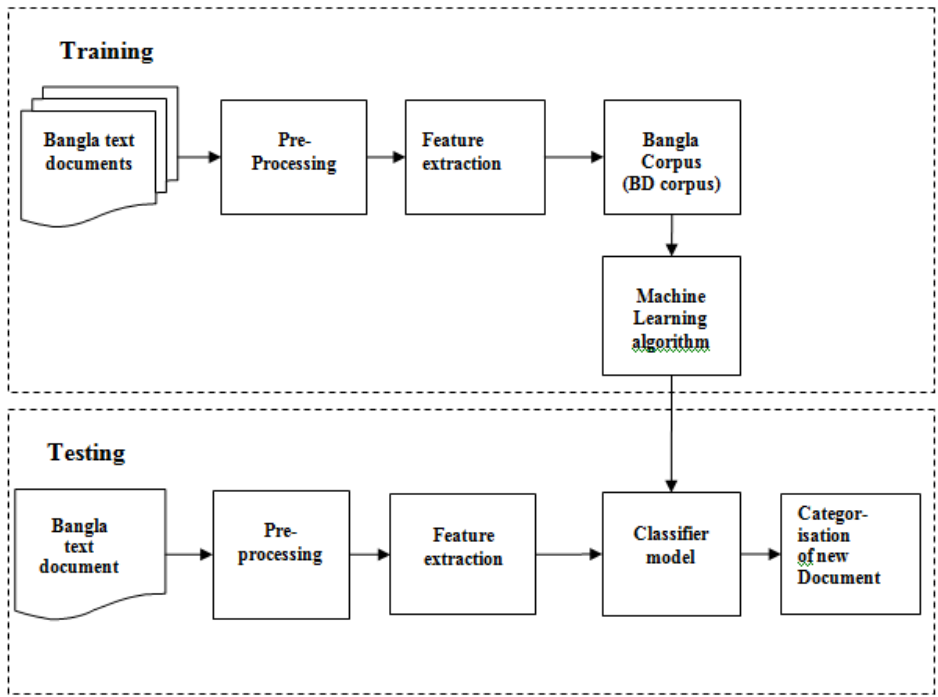

Figure 1. Bangla text classification process

## 4.1 Proposed Data Sets

Even though many corpora (document collections) in different language are available online for research purposes, in terms of Bangla language it is rare. Therefore, we have chosen to build our own corpus from various Bangla websites and named it BD corpus as presented in table 1. Each text document belongs to five categories: বাণিজ্য (business), খেলা (sports), স্বাস্থ্য (health), প্রযুক্তি (technology), শিক্ষা (education). The corpus includes 1000 documents with a total number of words being 22,218. It is also pointed out that without stop words, digits, punctuation and steaming words size is 18,190. All of these documents are taken from the Bangla news web sources such as *prothom-alo.com, online_dhaka.com, bdnews24.com, dailykalerkantha.com, bbc.co.uk/Bengali, ittefaque.com etc*, but all the documents are specific subject related.

International Journal of Artificial Intelligence & Applications (IJAIA), Vol. 5, No. 5, September 2014Table 1. Proposed BD corpus

| Category | Number of text documents |
|---|---|
| বাণিজ্য (business) | 160 |
| খেলা (sports) | 200 |
| স্বাস্থ্য (health) | 220 |
| প্রযুক্তি (technology) | 240 |
| শিক্ষা (education) | 180 |

### 4.2 Pre-processing

Choosing an appropriate representation of words in text documents is crucial to obtain good classification performance. Therefore, feature extraction or transforming the input data into the set of features is vital. It is expected that proper feature extraction will extract the relevant information from the input data, and due to the high dimensionality of feature sets, feature extraction can be performed to reduce the dimensionality of the feature space and improve the efficiency[33]. However, before extracting features from the documents, pre-processing is required. In pre-processing phase, we represent each original text document as "Bag of words". Then following operations are done on each document:

- **Tokenization:**

Tokenization is the process of breaking the sentences as well as the text file into word delimited by white space or tab or new line etc. Outcome of this tokenization phase is a set of word delimited by white space.

- **Digit Removal:**

A general Bengali text file may contain Bengali as well as English digits. But as meaningful Bengali words do not contain digits, we remove these digits by using their Unicode representation.

- **Punctuation Removal:**

We remove punctuation marks, special symbols (<,>, :,{,},[,],^,&,*,(,) , | etc.) from the Bengali text documents. Also, if a document contains excess use of spaces, tabs, shift, we remove them.

- **Stop words Removal:**

Stop words are the frequently occurring set of words which do not aggregate relevant information to the text classification task. Therefore, we have to remove these words from the text documents. We have compiled a list of Bengali language stop words from the dataset. This corpus contains around 364 words. Besides, there exist a lot of words having a single letter. Most of these Single-Letter-Words have little value. As a step of our processing, the stop-words need to be removed before further processing. So the Single-Letter-Words are removed in this phase. Some example of commonly occurring stop words are: এবং (*ebong*),

98

জন্য(*jonno*), আমি (*ami*), অনেকে (*oneke*), ইহা (*eha*), দিয়েছে (*dieche*), তারপর (*tarpor*), নিজে (*nije*),নাই (*nai*), ব্যাপারে (*bapare*) etc.

- **Stemming**

It is the process for reducing inflected (or sometimes derived) words to their stem, base or root form. E.g. বাংলাদেশে (*bangladeshe*)→বাংলাদেশ (*bangladesh*); আসনের (*ashoner*) → আসন (*ashon*) etc . An analysis of corpus was made and various possible noun suffixes were identified .Word stemming is an important pre-processing step because it works as dimension reduction.

## 4.3 Feature Extraction

After pre-processing phase we have documents with fewer numbers of words, and, therefore, extracting features from these words now become easier. The collection of words that are left in the document after all those steps are considered as a formal representation of the document, and we call the words in this collection 'terms'. This is our final text corpus. Various types of statistical approaches can be used to extract features from this text corpus. We use normalized (term frequency–inverse document frequency) TFIDF weighting with length normalization to extract the features from the document as this method performs better than many other methods [34]. A combination of term frequency and inverse document frequency called TFIDF is commonly used to represent term weight numerically. The weight for a term *i* in terms of TF-IDF is given by

$$w_i = \frac{\left(TF_i \times \log\left(\frac{N}{n_i}\right)\right)}{\sqrt{\sum_{i=1}^{n}\left(TF_i \times \log\left(\frac{N}{n_i}\right)\right)^2}} \tag{5}$$

where *N* is total number of documents and $n_i$ is document frequency of term *i*.

## 5. EXPERIMENT AND RESULT

### 5.1. Training and Testing Setup

In our experiment, after feature extraction and normalization, we have used the mentioned BD corpus as training datasets for obtaining models, and then, using these models we predict testing data sets. Algorithms experimented in this study are conducted with the help of Rapid Miner[35] and Java language. K fold strategy (with k=10) was followed to decide the number of training and testing examples. Nine folds, i.e. 900 examples were used as the training set to build the classifier and the remaining fold 100 examples were used to test the classifier for accuracy. Table 2 describe the K=10 fold strategy.

Table 2.The category and testing and training size of BD corpus

| Category | Total Sample | Training Sample | Testing Sample |
|---|---|---|---|
| বাণিজ্য (business) | 160 | 144 | 16 |
| খেলা (sports) | 200 | 180 | 20 |
| স্বাস্থ্য (health) | 220 | 198 | 22 |
| প্রযুক্তি (technology) | 240 | 216 | 24 |
| শিক্ষা (education) | 180 | 162 | 18 |

KNN experiments use the plain Euclidean distance as the similarity measure. On each pattern set containing a varying number of documents, different values of K from 1 to 10 are tested and the best results are reported.

SVM experiments use LIBSVM by Chung Chang and Chih-Jen Lin [36]. LIBSVM is integrated software for support vector classification, regression and distribution estimation. This LIBSVM is also an implementation of the SVM classifier algorithm that supports multiclass classification. According to our preliminary tests, the best results were achieved by the C-SVC method with the sigmoid kernel and we use this configuration.

One of the most well known decision tree algorithms is ID3 and its successor C4.5. For our experiments, we chose the C4.5 decision tree package since it has many nice features over its predecessor ID3 [37].

### 5.2. Performance Measures

Many evaluation metrics in text categorization are used. Our experiments adopt most commonly used performance measures, including the recall, precision, and F-measure ($F_1$). Given a test set A containing documents pre-labelled with category c and a prediction set B labelled with category c by the classifier, the recall (R) and precision (P) measures are defined by

$$R = \frac{|A \cap B|}{|A|} \quad (6)$$

and

$$P = \frac{|A \cap B|}{|B|} \quad (7)$$

respectively, where the norm |·| denotes the size of the document set [38] . It is a common practice to combine recall and precision in some way so that classifiers can be compared in terms of a single rating. Our Experiments used $F_1$ rating and it is defined as:

$$F_1 = \frac{2RP}{(R+P)} \quad (8)$$

Tests were performed on four classifiers: NB, KNN, DT (C4.5) and SVM. For each classifier, we measure the precession, recall and F-measure ($F_1$) experiment for each category. To evaluate the performance across categories, F-measure is averaged. There are two kinds of average value: micro average and macro average. We have used macro average. Table 3, 4, 5 and 6 describe the performance of four classifiers on the test corpus in terms of Precision, Recall and F-measure receptively.

Table 3. NB classifier results for five categories of BD corpus

| Category | Precision % | Recall % | F-measure % |
|---|---|---|---|
| বাণিজ্য (business) | 73.23 | 88.68 | 80.22 |
| (খলা (sports) | 91.59 | 74.42 | 82.12 |
| স্বাস্থ্য (health) | 85.77 | 93.29 | 89.37 |
| প্রযুক্তি (technology) | 96.43 | 87.34 | 91.66 |
| শিক্ষা (education) | 93.49 | 74.19 | 82.73 |
| | | Macro Average | 85.22 |

Table 4. KNN classifier results for five categories of BD corpus

| Category | Precision % | Recall % | F-measure % |
|---|---|---|---|
| বাণিজ্য (business) | 90.22 | 71.88 | 80.01 |
| খেলা (sports) | 73.19 | 61.33 | 66.74 |
| স্বাস্থ্য (health) | 93.37 | 67.67 | 78.47 |
| প্রযুক্তি (technology) | 63.63 | 86.25 | 73.23 |
| শিক্ষা (education) | 91.28 | 60.47 | 72.75 |
| | | Macro Average | 74.24 |

Table 5. DT (C4.5) classifier results for five categories of BD corpus

| Category | Precision % | Recall % | F-measure % |
|---|---|---|---|
| বাণিজ্য (business) | 73.93 | 88.29 | 80.47 |
| খেলা (sports) | 93.17 | 74.61 | 82.86 |
| স্বাস্থ্য (health) | 88.89 | 78.42 | 83.33 |
| প্রযুক্তি (technology) | 86.36 | 65.14 | 74.26 |
| শিক্ষা (education) | 91.58 | 74.78 | 82.33 |
| | | Macro Average | 80.65 |

Table 6. SVM classifier results for five categories of BD corpus

| Category | Precision % | Recall % | F-measure % |
|---|---|---|---|
| বাণিজ্য (business) | 84.14 | 88.58 | 84.14 |
| খেলা (sports) | 86.45 | 81.13 | 86.45 |
| স্বাস্থ্য (health) | 94.51 | 95.67 | 94.51 |
| প্রযুক্তি (technology) | 92.78 | 96.36 | 92.78 |
| শিক্ষা (education) | 87.80 | 88.11 | 87.80 |
| | | Macro Average | 89.14 |

After finding macro average F-measure ($F_1$) for four classifiers, it is observed that SVM achieved the highest average accuracy (89.14%), followed by NB with average accuracy of 85.22% and DT (C4.5) with average accuracy of 80.65%. KNN was the worst with average accuracy of 74.24%. Figure 2 shows the classifiers average performance.

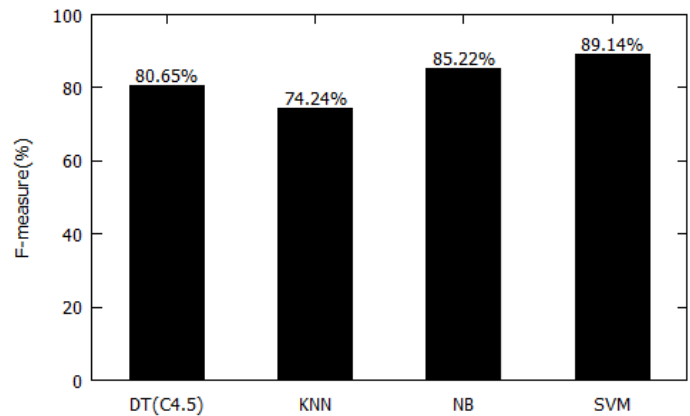

Figure 2. Average classification accuracy for four classifiers

### 5.3. Number of Training Examples and Accuracy

In this experiment, we are particularly interested in classifiers' learning abilities in various training sets. We train and test classifiers in 5 steps with each time increasing the input by 30 documents. Figure 3 depicts the number of training examples versus the accuracy in terms of average F-measure. Accuracy of the all classifiers was not very promising when only 30 documents were supplied as training data. However, the accuracy increases each time when classifiers are supplied with additional documents. In general, it can be surmised from the figure 2 that as long as we have sufficient number of text documents, all these four classifiers can learn reasonably well.

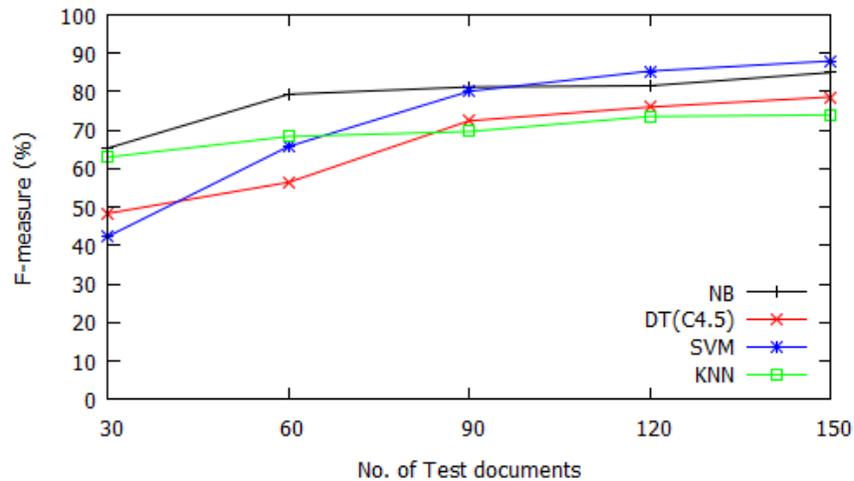

Figure3. Number of training documents versus accuracy

### 5.4. Training time

The amount of time required to train the classifiers is also a very important factor. Short training time is always demandable, but training phase for all classifiers usually takes time due to the nature of high dimensionality of the test datasets[39]. Figure 4 shows the average training time in seconds for the four text classifiers. This figure indicates that DT (C4.5) requires a significant amount of time for training compared to other three classifiers. DT (C4.5) is not scalable in high dimensional dataset, especially classifying large text documents. The classifier with the smallest training time is SVM, with around 9 seconds, whereas KNN and NB relatively have the similar training time, with 12.24 seconds and 14.15 seconds respectively.

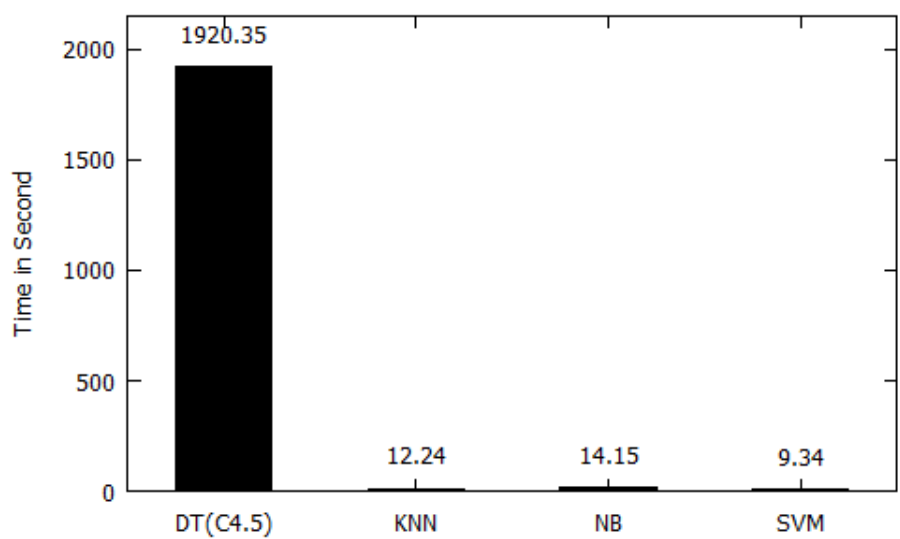

Figure 4. Average classifiers training time

## 6. CONCLUSIONS AND FUTURE DIRECTION

Developing efficient Bangla text classification is indispensable issue, as Bangla is a rich language with lot of diversity as well as Bangla based electronic documents both in online and offline have been emerged very quickly. In this article, we have revisited four state-of-the-art supervised machine learning algorithms including DT (C4.5), NB, KNN, and SVM and empirically compared their classification performances on Bangla text documents. In this respect, we have developed our own corpora, namely BD corpora, and implemented a tool for feature extraction and selection. The key findings of our experiments are summarized as follows:

- On small and well-organized training sets, KNN and NB are more capable than SVM and DT (C4.5) in categorization of documents. However, for large documents SVM is superior to other classifiers in text categorisation.
- Comparison of four classifiers in terms of training time indicates that all classifiers do not take the same learning time. DT (C4.5) takes more time than other three algorithms for training, whereas SVM is quick in learning.
- From the experiment, average F-measures indicate that SVM produces the best result followed by NB, DT (C4.5) and KNN. In general, if sufficient number of good training data is given, all the classifier produces satisfactory generalization performance on unseen test documents.
- SVM and KNN are sensitive to parameter settings. That is, different parameter produces a different result. Therefore, it is very crucial to define the best parameter settings.

The main future direction of our research is to examine the impact of other text representation schemes on text classification, such as some other term weighting schemes (e.g. tf-rf, tf-icf), latent semantic Indexing, and multi-word for text representation. For the sake of efficiency, it is also worth investigating further popular dimensionality reduction techniques including sophisticated term stemming and pruning.

## REFERENCES

[1]  F. Sebastiani, "Machine learning in automated text categorization," ACM computing surveys (CSUR), vol. 34, pp. 1-47, 2002.
[2]  M. Sugiyama and M. Kawanabe, Machine learning in non-stationary environments: Introduction to covariate shift adaptation: MIT Press, 2012.
[3]  A. Lopes, R. Pinho, F. V. Paulovich, and R. Minghim, "Visual text mining using association rules," Computers & Graphics, vol. 31, pp. 316-326, 2007.
[4]  T. Denoeux, "A k-nearest neighbor classification rule based on Dempster-Shafer theory," Systems, Man and Cybernetics, IEEE Transactions on, vol. 25, pp. 804-813, 1995.
[5]  J. R. Quinlan, "Induction of decision trees," Machine learning, vol. 1, pp. 81-106, 1986.
[6]  J. Chen, H. Huang, S. Tian, and Y. Qu, "Feature selection for text classification with Naïve Bayes," Expert Systems with Applications, vol. 36, pp. 5432-5435, 2009.
[7]  C. Cortes and V. Vapnik, "Support-vector networks," Machine learning, vol. 20, pp. 273-297, 1995.
[8]  P. F. Brown, P. V. Desouza, R. L. Mercer, V. J. D. Pietra, and J. C. Lai, "Class-based n-gram models of natural language," Computational linguistics, vol. 18, pp. 467-479, 1992.
[9]  M. Islam, "Research on Bangla language processing in Bangladesh: progress and challenges," 8th ILDC, Dhaka, Bangladesh (June 2009), 2009.
[10] P. S. Jacobs, Text-based intelligent systems: Current research and practice in information extraction and retrieval: Psychology Press, 2014.
[11] A. K. Mandal, M. D. Hossain, and M. Nadim, "Developing an efficient search suggestion generator, ignoring spelling error for high speed data retrieval using Double Metaphone Algorithm," in Computer and Information Technology (ICCIT), 2010 13th International Conference on, 2010, pp. 317-320.
[12] M. Mansur, "Analysis of n-gram based text categorization for bangla in a newspaper corpus," BRAC University, 2006.
[13] B. Agarwal and N. Mittal, "Text Classification Using Machine Learning Methods-A Survey," in Proceedings of the Second International Conference on Soft Computing for Problem Solving (SocProS 2012), December 28-30, 2012, 2014, pp. 701-709.
[14] F. Ciravegna, L. Gilardoni, A. Lavelli, S. Mazza, W. J. Black, M. Ferraro, et al., "Flexible text classification for financial applications: the FACILE system," in ECAI, 2000, pp. 696-700.
[15] T. Zagibalov and J. Carroll, "Automatic seed word selection for unsupervised sentiment classification of Chinese text," in Proceedings of the 22nd International Conference on Computational Linguistics-Volume 1, 2008, pp. 1073-1080.
[16] A. Moh'd Mesleh, "Support vector machines based Arabic language text classification system: feature selection comparative study," in Advances in Computer and Information Sciences and Engineering, ed: Springer, 2008, pp. 11-16.
[17] K. Rajan, V. Ramalingam, M. Ganesan, S. Palanivel, and B. Palaniappan, "Automatic classification of Tamil documents using vector space model and artificial neural network," Expert Systems with Applications, vol. 36, pp. 10914-10918, 2009.
[18] N. a. V. Gupta, "Algorithm for Punjabi Text Classification," International Journal of Computer Applications, vol. 37, pp. 30-35, 2012.
[19] A. S. Patil and B. Pawar, "Automated classification of web sites using Naive Bayesian algorithm," in Proceedings of the International MultiConference of Engineers and Computer Scientists, 2012, pp. 14-16.
[20] L. Jiang, Z. Cai, H. Zhang, and D. Wang, "Naive Bayes text classifiers: a locally weighted learning approach," Journal of Experimental & Theoretical Artificial Intelligence, vol. 25, pp. 273-286, 2013.
[21] Q. Yuan, G. Cong, and N. M. Thalmann, "Enhancing naive bayes with various smoothing methods for short text classification," in Proceedings of the 21st international conference companion on World Wide Web, 2012, pp. 645-646.
[22] V. Bijalwan, V. Kumar, P. Kumari, and J. Pascual, "KNN based Machine Learning Approach for Text and Document Mining," International Journal of Database Theory and Application, vol. 7, pp. 61-70, 2014.
[23] B. Li, S. Yu, and Q. Lu, "An improved k-nearest neighbor algorithm for text categorization," arXiv preprint cs/0306099, 2003.
[24] N. Suguna and K. Thanushkodi, "An improved K-nearest neighbor classification using Genetic Algorithm," International Journal of Computer Science Issues, vol. 7, pp. 18-21, 2010.

International Journal of Artificial Intelligence & Applications (IJAIA), Vol. 5, No. 5, September 2014

[25] X.-L. Liu, S. Ding, H. Zhu, and L. Zhang, "Appropriateness in applying SVMs to text classification," Comput Eng Sci, vol. 32, pp. 106-108, 2010.
[26] T. Zakzouk and H. Mathkour, "Text Classifiers for Cricket Sports News," in proceedings of International Conference on Telecommunications Technology and Applications ICTTA, 2011, pp. 196-201.
[27] L. Zhijie, L. Xueqiang, L. Kun, and S. Shuicai, "Study on SVM Compared with the other Text Classification Methods," in Education Technology and Computer Science (ETCS), 2010 Second International Workshop on, 2010, pp. 219-222.
[28] A. S. Babu and P. Kumar, "Comparing Neural Network Approach with N-Gram Approach for Text Categorization," Int J Comput Sci Engin, vol. 2, pp. 80-83, 2010.
[29] C. Vens, J. Struyf, L. Schietgat, S. Džeroski, and H. Blockeel, "Decision trees for hierarchical multi-label classification," Machine Learning, vol. 73, pp. 185-214, 2008.
[30] M. Ghiassi, J. Skinner, and D. Zimbra, "Twitter brand sentiment analysis: A hybrid system using n-gram analysis and dynamic artificial neural network," Expert Systems with Applications: An International Journal, vol. 40, pp. 6266-6282, 2013.
[31] Y. Yang and X. Liu, "A re-examination of text categorization methods," in Proceedings of the 22nd annual international ACM SIGIR conference on Research and development in information retrieval, 1999, pp. 42-49.
[32] Y. H. Li and A. K. Jain, "Classification of text documents," The Computer Journal, vol. 41, pp. 537-546, 1998.
[33] G. Forman and E. Kirshenbaum, "Extremely fast text feature extraction for classification and indexing," in Proceedings of the 17th ACM conference on Information and knowledge management, 2008, pp. 1221-1230.
[34] H. Joho and M. Sanderson, "Document frequency and term specificity," in Large Scale Semantic Access to Content (Text, Image, Video, and Sound), 2007, pp. 350-359.
[35] G. Ertek, D. Tapucu, and İ. Arın, "Text mining with rapidminer," RapidMiner: Data Mining Use Cases and Business Analytics Applications, p. 241, 2013.
[36] C.-C. Chang and C.-J. Lin, "LIBSVM: a library for support vector machines," ACM Transactions on Intelligent Systems and Technology (TIST), vol. 2, p. 27, 2011.
[37] D. L. Olson and D. Delen, Advanced data mining techniques: Springer, 2008.
[38] C. J. V. Rijsbergen, Information Retrieval: Butterworth-Heinemann, 1979.
[39] F. Alam, F. Bappee, M. R. Rabbani, and M. M. Islam, "An Optimized Formulation of Decision Tree Classifier," in Advances in Computing, Communication, and Control. vol. 361, S. Unnikrishnan, S. Surve, and D. Bhoir, Eds., ed: Springer Berlin Heidelberg, 2013, pp. 105-118.

**AUTHORS**

**ASHIS KUMAR MANDAL** received B.Sc. (Eng.) in CSE from Sahajalal University of Science and Technology, Bangladesh. He is currently doing M.Sc in Computer Science at University Malaysia Pahang, Malaysia. He has been joined as a lecturer in the Department of Computer Engineering, Hajee Md. Danesh Science and Technology University, Bangladesh. Currently he is an Assistant Professor. His research interests include data mining, machine learning, optimization, network security.

**RIKTA SEN** received B.Sc. (Eng.) in Electronic and Telecommunication Engineering from Rajshahi University of Engineering and Technology, Bangladesh. She is currently doing M.Sc in EEE at Rajshahi University of Engineering and Technology. Her research areas are data mining, Wireless Sensor Networks, Optical fiber and Mobile communication.